\documentclass[
]{ceurart}

\usepackage{listings}

\usepackage{algorithm}
\usepackage{algpseudocode}

\usepackage{subcaption}
\usepackage[all]{nowidow}
\begin{document}

\copyrightyear{2022}
\copyrightclause{Copyright for this paper by its authors.
  Use permitted under Creative Commons License Attribution 4.0
  International (CC BY 4.0).}

\conference{LDAC'25: Linked Data in Architecture and Construction, July 09--11, 2025, Porto, Portugal}

\title{Representing Normative Regulations in OWL DL for Automated Compliance Checking Supported by Text Annotation}


\author[1]{Ildar Baimuratov}[%
orcid=0000-0002-6573-131X,
email=ildar.baimuratov@l3s.de
]
\address[1]{L3S Research Center, Leibniz University Hannover, Germany}

\author[2]{Denis Turygin}[%
orcid=0009-0003-5502-4077,
email=turygindenis7@gmail.com
]
\address[2]{ITMO University, St. Petersburg, Russia}




\begin{abstract}
Compliance checking is the process of determining whether a regulated entity adheres to these regulations. Currently, compliance checking is predominantly manual, requiring significant time and highly skilled experts, while still being prone to errors caused by the human factor. Various approaches have been explored to automate compliance checking, however, representing regulations in OWL DL language which enables compliance checking through OWL reasoning has not been adopted. In this work, we propose an annotation schema and an algorithm that transforms text annotations into machine-interpretable OWL DL code. The proposed approach is validated through a proof-of-concept implementation applied to examples from the building construction domain.
\end{abstract}

\begin{keywords}
  Normative Regulations \sep
  Text Annotation \sep
  OWL DL\sep
  Reasoning \sep
  Compliance checking
\end{keywords}

\maketitle

\section{Introduction}

Normative regulations govern business processes, industry, law, and various other domains. The process of verifying whether a regulated entity meets these regulations is known as compliance checking. Currently, this process is predominantly manual, requiring significant time and expertise while remaining prone to human error. For instance, in building construction, a single compliance review cycle can take several weeks, and multiple review cycles may be necessary due to design modifications. Non-compliance with building regulations can result in fines, penalties, or even criminal prosecution. Moreover, studies have shown significant discrepancies in manual code reviews, with different plan review departments often reaching inconsistent conclusions when evaluating the same set of plans \cite{Zhang2017Jan}. Additionally, the redundancy of building codes contributes to inefficiencies and increases the likelihood of errors during the compliance checking process \cite{fitkau2024ontology} \cite{vanbim}.

Thus, there is a clear need to automate compliance checking. However, normative regulations are typically presented in human-readable formats, making them incompatible with software processing. Compared to manual compliance checking, automated compliance checking (ACC) is expected to improve efficiency by reducing time, costs, and errors. However, existing compliance checking systems, such as the Solibri\footnote{\url{https://www.solibri.com/}} building model checker, rely on manually created, hard-coded, proprietary rules to represent normative regulations. While effective for a specific set of regulations within a given timeframe, this rigid approach requires significant effort to adapt to different regulatory codes and maintain over time. Machine learning (ML), particularly large language models (LLMs), offers potential support for ACC. However, the trustworthiness of ML models remains questionable due to issues such as hallucinations, lack of transparency, and limited reproducibility --- critical factors in responsible domains like building construction.

In contrast, symbolic reasoning is inherently accurate, reproducible, and explainable. In recent years researchers have investigated the use of Semantic Web technologies, such as RDF, OWL, SPARQL, SWRL and SHACL for compliance checking. However, to the best of our knowledge, no study has applied OWL DL\footnote{\url{https://www.w3.org/TR/2012/REC-owl2-syntax-20121211/}} reasoning for ACC. Among the most relevant works, \citet{fitkau2024ontology} modeled part of building regulations using OWL DL but processed them solely with DL querying. \citet{nuyts2024comparative} employed OWL DL to check information availability, while compliance constraints were modeled using SPARQL, SWRL, and SHACL. However, query-based approaches, such as SHACL and DL Query, have several limitations, including the lack of established semantics; the absence of consistency checking; the inability to trace violations back to their sources; non-completely human-readable syntax; and dependence on the RDF graph structure. In contrast, OWL DL reasoning offers several advantages, such as a standardized, human-readable (Manchester) syntax; semantics grounded in decidable description logics (DLs); independence from data complexity; explanations that ensure traceability; and identifying redundant or inconsistent regulations. The primary challenge that hinders researchers from using OWL is the open-world assumption (OWA). However, we argue that OWL DL is expressive enough to produce the same results as closed-world reasoning, provided that the data is modeled correctly.


Converting textual regulations into machine-readable formats such as OWL DL remains a challenging task. Various approaches have been explored to facilitate the formalization of regulations, including NLP techniques to generate Prolog clauses \cite{Zhang2017Jan} or SHACL shapes \cite{donkers2024converting}, deep learning for LegalRuleML \cite{fuchsa2024transformer}, and Large Language Models (LLMs) for SPARQL \cite{chen2024automated}. However, since modeling regulations in OWL DL has been largely unexplored, there are no studies on translating regulation texts to OWL DL.
To address these gaps, we propose a text annotation schema and an algorithm for automatically converting annotated regulations into OWL DL code. The annotation schema facilitates the alignment of text with the regulations' semantics, making it accessible to domain experts. It also leverages existing annotation tools, removing the need for custom formalization interfaces, and paves the way for future integration with machine learning models to support the annotation process.

The contributions of this research include:
\begin{itemize}
    \item An approach for representing normative regulations in OWL DL that enables ACC through OWL reasoning. 
    \item A text annotation schema and an algorithm for automatically converting annotated regulations into OWL DL code.
\end{itemize}
The proposed approach is validated through a proof-of-concept implementation applied to examples from the building construction domain.



The paper is organized as follows: \autoref{sec:rel} reviews the relevant research, \autoref{sec:annotation} describes the proposed text annotation schema, \autoref{sec:req2owl} presents the algorithm for converting regulations into OWL DL code, and \autoref{sec:proof} demonstrates the proof of concept.
\section{Related Work}
\label{sec:rel}

In this section, we review research relevant to our work, focusing on approaches for machine-readable representation of regulations and the streamlining of their formalization. Additionally, we argue in favor of using OWL DL over SHACL.

\subsection{Regulation representation}

Normative regulations are studied across various disciplines, including law, legal reasoning, deontic logic, and artificial intelligence. Researchers have explored the formalization of these regulations to facilitate ACC. In this subsection, we focus specifically on the manual conversion of the regulations.

\citet{Hashmi} proposed an abstract regulatory compliance framework for business processes based on deontic logic. Among machine-readable representation, LegalRuleML \cite{LegalRuleML} extended the syntax of RuleML\footnote{\url{https://www.ruleml.org/}} with concepts and features specific to legal norms. However, LegalRuleML provides only mechanisms to capture and represent different interpretations of legal norms, without relying on any specific logical framework. \citet{Gandon} explored the application of Semantic Web frameworks to the formalization and processing of normative regulations. They built upon the LegalRuleML model, incorporating notions of regulatory compliance from \cite{Hashmi}. Their approach modeled states of affairs as named graphs and utilizes SPARQL for querying. \citet{legal} also utilized SPARQL to reason over legislation within a description logic (DL) framework. \citet{Lam} presented a method for transforming legal norms from LegalRuleML into a variant of Modal Defeasible Logic. This transformation was implemented as an extension to the DL reasoner SPINdle.


In the building construction domain, the use of Semantic Web technologies has also been evaluated. \citet{yurchyshyna2009} represented constraints with SPARQL queries. \citet{pauwels2011} utilized N3 logic. The SWRL language was used to encode rules in \cite{beach2015}, \cite{Fahad2016Oct} and \cite{Shi2017Jan}. \citet{fitkau2024ontology} developed a Fire Safety Ontology using SPARQL, DL Query and SWRL. The use of SHACL for compliance checking has been evaluated in \cite{stolk2020}, \cite{kovacs2021bim},\cite{zentgraf2022multi}, \cite{nuyts2023validation},\cite{donkers2024converting} and \cite{patlakas2024semantic}. The use of OWL was previously demonstrated in \cite{nuyts2024comparative}, however, the authors applied it solely to model information availability constraints, while compliance constraints were modeled using SPARQL, SWRL, or SHACL.

\subsection{OWL vs SHACL}

Researchers have raised concerns that OWL, based on the Open World Assumption (OWA), may not effectively handle constraints for compliance checking. As a result, recent studies have predominantly used SHACL, which relies on the Closed World Assumption (CWA). In this subsection, we advocate for using OWL DL to model regulations.

First, we compare OWL DL and SHACL in terms of their expressive power and computational decidability by relating them to the common framework of description logics (DLs). \citet{leinberger2020} map SHACL shapes to DLs, enabling shape containment to be addressed through description logic reasoning. They conclude that the corresponding description logic for SHACL shapes is ALCOIQ($\circ$), which is undecidable for infinite models. A restricted fragment of SHACL, limiting path concatenation, corresponds to the description logic SROIQ, which underpins OWL DL. \citet{bogaerts2022shacl} explore the relationship between SHACL and OWL, arguing that SHACL is, in fact, a description logic. However, they point out that other tasks typically studied in DLs, such as consistency checking, are undecidable in SHACL, except for finite model checking.

Second, although OWL is based on OWA, it is still expressive enough to support closed-world reasoning. To achieve this, missing information and disjointness must be explicitly defined. For instance, if an individual $a$ has only one relation $R(a,b)$ (in DL notation), this fact must be explicitly declared in the ontology at the time of compliance checking as $a \in \forall R.\{b\}$. Alternatively, if $a$ has no relations at all, and there exists a relation $R$ in the ontology, it must be formulated as $a \in \forall R.\bot$. Software tools like owlready2 \cite{Owlready} provide methods to algorithmically apply local closure to specific individuals, classes, or even entire ontologies. Once closed, these ontologies can be processed by OWL reasoners in the usual manner. Therefore, the OWA is more relevant to data modeling than to regulation modeling, and the former does not require additional effort from users.

Finally, OWL offers better human-computer interaction. \citet{ahmetaj2021reasoning} emphasize that in SHACL, validation reports provide limited information, mainly identifying the node and indicating constraint violations. In contrast, OWL reasoners offer detailed explanations of classifications and inconsistencies, allowing users to trace them back to their sources. Additionally, OWL DL is supported by the Manchester syntax, which is more human-readable compared to the serialization format available for SHACL. 


\subsection{Streamlining regulation formalization}

Converting textual rules into machine-readable formats is a challenging task. Researchers have explored the potential of Natural Language Processing (NLP) and Machine Learning (ML) techniques to translate natural language regulatory texts into specific representation formats. \citet{Hjelseth2011} introduced an annotation format for normative texts called RASE, although the study does not address the conversion of these annotations into an executable format. \citet{Zhang2017Jan} proposed an ACC system that uses NLP techniques to automatically extract normative information from documents and convert it into logical rules, opting for first-order logic and its implementation in B-Prolog. \citet{donkers2024converting} leveraged the linguistic structure of sentences to automatically generate SHACL representations using predefined templates. Recent studies also explore automating regulation formalization with LegalRuleML via deep learning \cite{fuchsa2024transformer} or with SPARQL using LLMs \cite{chen2024automated}. To address the lack of resources for ML, \citet{hettiarachchi2025code} introduced CODE-ACCORD, a dataset of sentences from the building regulations of England and Finland, annotated with a custom set of entities and relations not grounded in any formal framework.

In contrast to these approaches, we propose an annotation schema that directly aligns with OWL DL syntax, ensuring that the translation of annotated regulations is both transparent and human-controllable. The intermediate annotation step facilitates the alignment of the text with the regulations' semantics, making it more accessible to domain experts, and leverages existing annotation tools, eliminating the need for custom formalization interfaces. Finally, machine learning models can be trained on these annotations to further assist in the annotation process.
\section{Annotation Schema}
\label{sec:annotation}

In this section, we introduce a schema for annotating regulatory text to streamline its translation into OWL DL code. This schema facilitates the alignment between textual content and regulatory semantics, making it accessible to domain experts. By leveraging existing annotation tools, it eliminates the need for custom formalization interfaces. The annotation schema includes three layers: Terms, Semantic Types, and Semantic Roles. Each layer includes both span-based tags and arrows connecting them. For clarity, annotation tags are written in \textit{italics} and OWL expressions are written in \texttt{teletype}. The annotation schema is illustrated with two examples: one qualitative regulation and one quantitative.

\paragraph{Terms.}
The first annotation layer focuses on domain terms. This layer is essential for aligning the formalized regulations with the domain data for validation and for enabling efficient search and filtering of regulations. This layer utilizes domain vocabularies or ontologies as tags. For instance, in the building construction domain, Building Information Models (BIM) \cite{BIM} are defined using the Industry Foundation Classes (IFC)\footnote{\url{https://technical.buildingsmart.org/standards/ifc}}. Its OWL serialization, ifcOWL \cite{ifcOWL}, can be employed as values within the Term layer.


\paragraph{Semantic Types.}
The Semantic Types layer includes tags that correspond to syntactic elements of the OWL DL language, enabling the construction of class restrictions. We primarily use Manchester syntax\footnote{\url{https://www.w3.org/TR/owl2-manchester-syntax/}} for tag names, though some are slightly modified for better readability by domain experts. For example, \texttt{owl:ObjectProperty} is represented by the tag \textit{Relation}, and \texttt{owl:DataProperty} corresponds to \textit{Property}. The mapping of these tags to OWL is provided in \autoref{tab:types}. Additionally, the Semantic Types layer includes the specific arrows \textit{Domain}, \textit{Range} and \textit{Of}. Their possible starting and ending tags are provided in \autoref{tab:type_arrows}.



If a Term tag appears on the same tokens as a Semantic Type tag, the latter might seem redundant. However, our goal is to provide a comprehensive representation of the regulation's semantics at the Semantic Type layer, independent of the specific vocabulary used in the Term layer. This approach ensures that Semantic Types can be reused across different domain models (or even without any). As a result, raw OWL code may contain multiple instances of the same concept, reflecting its various spellings and synonyms. However, since our aim is not to create a single ontology for all regulations but rather a set of individual OWL DL programs, each representing a regulation, we do not need to align these variations.

\begin{table}[htb!]
\begin{minipage}[c]{.3\linewidth}
  \caption{Mapping of Semantic Types to the OWL DL\\syntax}
  \label{tab:types}
  \begin{tabular}{cl}
    \toprule
    Tag & OWL \\
    \midrule
    Literal & literal \\
    Class & owl:Class \\
    Not & owl:complementOf \\
    Or & owl:unionOf \\
    Relation & owl:ObjectProperty \\
    Property & owl:DataProperty \\
    Some & owl:someValuesFrom \\
    Only & owl:allValuesFrom \\
    Number & \parbox[t]{3.5cm}{owl:maxCardinality, owl:minCardinality, owl:cardinality} \\
    Comparison & \parbox[t]{3.5cm}{xsd:minInclusive, xsd:minExclusive, xsd:maxInclusive, xsd:maxExclusive} \\
    \bottomrule
  \end{tabular}
\end{minipage}%
\hfill
\begin{minipage}[c]{.55\linewidth}
  \caption{Arrows connecting Semantic Type tags}
  \label{tab:type_arrows}
  \begin{tabular}{cp{3cm}p{3cm}}
    \toprule
    Arrow & Start & End \\
    \midrule
    Domain & Relation, Property & Class, Relation \\
    Range & Relation & Class, Relation, Property \\
    & Property & Literal \\
    Of & Not, Or & Class, Relation, Property \\
    & Comparison & Literal \\
    & Some, Only, Number & Relation, Property \\
    \bottomrule
  \end{tabular}
\end{minipage} 
\end{table}


\paragraph{Semantic Roles.}
Semantic Roles are essential for constructing axioms in the OWL DL language. OWL axioms are statements that are asserted to be true within the domain of interest. In OWL, two types of assertions are possible: one about the relation between individuals, and the other about the membership of an individual or class in a class. In the context of normative regulations, we focus on asserting relationships between restricted classes (annotated with Semantic Types), so only the relation \texttt{owl:SubClassOf}, or General Class Inclusion (GCI) is needed.
To annotate this relation, two semantic roles are introduced: 1) \textit{Subject}, which represents the subclass, 2) and \textit{Requirement}, which represents the superclass. In other words, the \textit{Subject} role denotes the OWL class to which the regulation applies, while the \textit{Requirement} role represents the class containing the imposed regulation. The arrow between \textit{Subject} and \textit{Requirement}, corresponding to the \texttt{owl:SubClassOf} relation is annotated with the tag \textit{To}, resulting in the triple \textit{<Subject, To, Requirement>}.

\paragraph{Linguistic arrows.}
In addition to arrows that represent the semantic relationships between different tags, the proposed annotation schema also includes linguistic arrows, which only connect separate pieces of text related to the same tag. There are three linguistic arrows with distinct properties, which are described as follows:
\begin{itemize}
    \item \textit{Concatenation}: $a\rightarrow b\Rightarrow ab$,
    \item \textit{Distribution}: $a\rightarrow b, a\rightarrow c\Rightarrow ab, ac$,
    \item \textit{Self-Distribution}: $a\rightarrow b\Rightarrow a, ab$,
\end{itemize}
where $a$, $b$, and $c$ are pieces of text and $ab$ or $ac$ represent $a$ and $b$ or $a$ and $c$ respectively concatenated. \textit{Self-distribution} can be considered as \textit{Distribution} in which one of the strings is empty:
\begin{equation*}
\label{eq:self_distr_eq}
    a\rightarrow 0, a\rightarrow b\Rightarrow a, ab.
\end{equation*}
However, since not all tools allow annotating an empty line in the text, we introduce \textit{Self-distribution} as a separate arrow.



\paragraph{Annotation Interface.}
To annotate regulations following the proposed schema, we use the INCEpTION tool \cite{Inception}. One of the key advantages of this tool is its ability to import ontologies, which can then be used as annotation tags. In our use-case, the imported ifcOWL ontology serves as the tagset for the Term layer, and tagsets for other layers are created manually. Additionally, this tool supports the integration of ML-based recommenders to streamline the annotation process. The original regulations are imported into INCEpTION in TXT format, and the annotations are exported in WebAnno TSV v3.3. \nameref{ex1} and \nameref{ex2} illustrate annotated regulations, one qualitative and the other quantitative.

\begin{figure*}[htb!]
	\centering
	\includegraphics[width=0.9\linewidth]{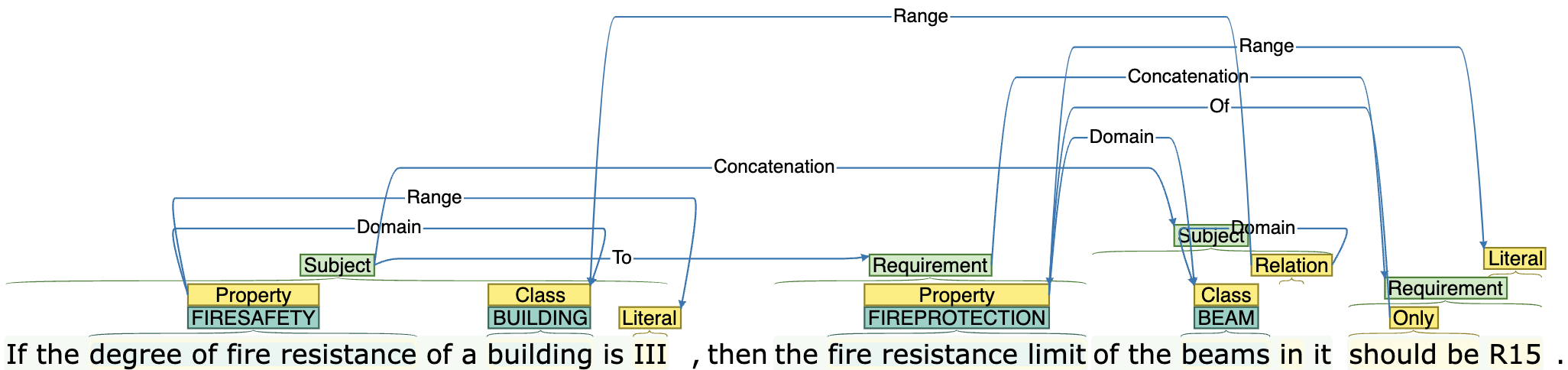}
    \caption{Annotation of \nameref{ex1} in INCEpTION}
    \label{fig:ex}
\end{figure*}

\paragraph{Example 1.}
\label{ex1}
As an example, consider a qualitative regulation from the building construction domain with the text ``If the degree of fire resistance of a building is III, then the fire resistance limit of the beams in it should be R15''. In this regulation, we annotate the terms \textit{ifcBuilding}, \textit{ifcBeam}, \textit{FIRESAFETY}, and \textit{FIREPROTECTION}. The first two are annotated with the Semantic Type \textit{Class}, while the latter two are annotated as \textit{Property}. Additionally, ``III'' and ``R15'' are annotated as \textit{Literal}, the preposition ``in'' is labeled as a \textit{Relation}, and ``should be'' is annotated with the tag \textit{Only}. Finally, the phrases ``If the degree of fire resistance of a building is III'' and ``the beams in it'' are connected with a \textit{Concatenation} arrow and form together a subject of the regulation. Similarly, the phrases ``then the fire resistance limit'' and ``should be R15'' are also connected with a \textit{Concatenation} arrow and identified as the regulation's requirement. \autoref{fig:ex} shows the annotation of this regulation within the INCEpTION interface.

\begin{figure*}[htb!]
	\centering
	\includegraphics[width=0.9\linewidth]{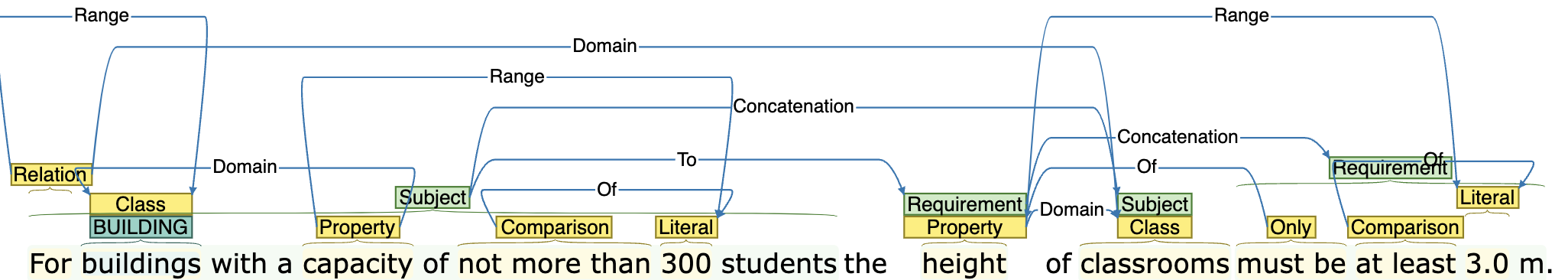}
    \caption{Annotation of \nameref{ex2} in INCEpTION}
    \label{fig:ex2}
\end{figure*}

\paragraph{Example 2.}
\label{ex2}
As a quantitative requirement, consider the following: ``For buildings with a capacity of not more than 300 students the height of classrooms must be at least 3.0 m''. Compared to \nameref{ex1}, the subject of this regulation contains a chain consisting of the relation ``for'' and the property ``capacity'', as well as the constrained data type ``not more than 300''. Its requirement also includes the constrained data type ``at least 3.0''. \autoref{fig:ex2} illustrates the annotation of this regulation.

\section{Transforming Annotations into OWL DL}
\label{sec:req2owl}

In this section, we present our algorithm for transforming annotated regulations into OWL DL code, which enables the classification of domain objects into regulation subjects and the subsequent checking of their compliance with the imposed requirements with OWL reasoning.
The algorithm is divided into several steps: 1) data preprocessing, 2) generating elementary OWL entities, 3) aligning domain terms with semantic types, 4) vocabulary-based mapping of numbers and comparisons, 5) constructing class restrictions, and finally, 6) constructing axioms. The algorithm is recursive, allowing it to process annotations of any complexity.

\paragraph{Preprocessing.} The algorithm takes as input a TSV file exported from INCEpTION, which contains the annotated regulation in a tabular format. In this table, each row  corresponds to a word-based token, indexed by its order of appearance in the text, and each column represents a distinct annotation layer.
Before generating OWL code, the data undergoes preprocessing in two stages. First, linguistic arrows are processed, so that each table row corresponds to exactly one tag. New tags derived from these operations are added to the table, while the old ones are removed. Second, tokens are grouped and extracted into three separate tables, each corresponding to an annotation layer: the table of Terms $Tr$, the table of Semantic Types $ST$, and the table of Semantic Roles $SR$. Additionally, we use lowercase letters to denote single elements, while uppercase denotes sets. For example, $tr$ refers to a single row from the table of Terms $Tr$. \autoref{tab:ex1_terms}, \autoref{tab:ex1_types}, and \autoref{tab:ex1_roles} present the tables for \nameref{ex1} and \autoref{tab:ex2_terms}, \autoref{tab:ex2_types}, and \autoref{tab:ex2_roles} correspond to \nameref{ex2}.


\begin{table}[htb!]
\caption{Layer tables from \nameref{ex1}}
\begin{minipage}{0.475\linewidth}
\begin{subtable}[htb!]{\linewidth}
\centering
\caption{Terms}
\label{tab:ex1_terms}
\begin{tabular}{cp{2cm}c}
    \toprule
    Index & Token & Tag \\
    \midrule
        1 & beams & BEAM \\
        2 & building & BUILDING \\
        9 & fire resistance limit & FIREPROTECTION \\
        8 & degree of fire resistance & FIRESAFETY \\
    \bottomrule
    \end{tabular}
\end{subtable}%
\hfill
\begin{subtable}[htb!]{\linewidth}
\centering
    \caption{Semantic Roles}
    \label{tab:ex1_roles}
\begin{tabular}{>{\centering\arraybackslash}p{0.5cm} >{\centering\arraybackslash}p{3cm} >{\centering\arraybackslash}p{1cm} >{\centering\arraybackslash}p{1cm}}
    \toprule
    Index & Token & Tag & To \\
    \midrule
    0 & If the degree of fire resistance of a building is III of the beams in it & Subject & 1 \\
    1 & the fire resistance limit should be R15 & Requirement & - \\
    \bottomrule
\end{tabular}
\end{subtable}
\end{minipage}
\begin{minipage}{0.475\linewidth}
\begin{subtable}[htb!]{\linewidth}
\centering
\caption{Semantic types}
\label{tab:ex1_types}
\begin{tabular}{>{\centering\arraybackslash}p{0.5cm} >{\centering\arraybackslash}p{1.5cm} >{\centering\arraybackslash}p{0.7cm} >{\centering\arraybackslash}p{0.7cm} >{\centering\arraybackslash}p{0.7cm} >{\centering\arraybackslash}p{0.7cm}}
    \toprule
    Index & Token & Tag & Domain & Range & Of \\
    \midrule
    1-20 & beams & Class & - & - & - \\
    1-9 & building & Class & - & - & - \\
    1-11 & III & Literal & - & - & - \\
    1-25 & R15 & Literal & - & - & - \\
    5 & degree of fire resistance & Property & 1-9 & 1-11 & - \\
    6 & fire resistance limit & Property & 1-20 & 1-25 & - \\
    1-21 & in & Relation & 1-20 & 1-9 & - \\
    7 & should be & Only & - & - & 6 \\
    \bottomrule
\end{tabular}
\end{subtable}
\end{minipage}
\end{table}

\begin{table}[htb!]
\caption{Layer tables from \nameref{ex2}}
\begin{minipage}{0.475\linewidth}
\begin{subtable}[htb!]{\linewidth}
\centering
\caption{Terms}
\label{tab:ex2_terms}
\begin{tabular}{cccc}
    \toprule
    Index & Token & Tag \\
    \midrule
        5 & buildings & BUILDING \\
    \bottomrule
\end{tabular}
\end{subtable}
\begin{subtable}[htb!]{\linewidth}
\centering
    \caption{Semantic Roles}
    \label{tab:ex2_roles}
\begin{tabular}{>{\centering\arraybackslash}p{0.5cm} >{\centering\arraybackslash}p{3cm} >{\centering\arraybackslash}p{1cm} >{\centering\arraybackslash}p{1cm}}
    \toprule
    Index & Token & Tag & To \\
    \midrule
    0 & For buildings with a capacity of not more than 300 students classrooms & Subject & 1 \\
    1 & height must be at least 3.0 m & Requirement & - \\
    \bottomrule
\end{tabular}
\end{subtable}
\end{minipage}%
\hfill
\begin{minipage}{0.475\linewidth}
\begin{subtable}[htb!]{\linewidth}
\centering
\caption{Semantic types}
\label{tab:ex2_types}
\begin{tabular}{>{\centering\arraybackslash}p{0.5cm} >{\centering\arraybackslash}p{1.5cm} >{\centering\arraybackslash}p{0.7cm} >{\centering\arraybackslash}p{0.7cm} >{\centering\arraybackslash}p{0.7cm} >{\centering\arraybackslash}p{0.7cm}}
    \toprule
    Index & Token & Tag & Domain & Range & Of \\
    \midrule
    1-15 & classrooms & Class & - & - & - \\
    1-2 & buildings & Class & - & - & - \\
    3 & not more than & Comparison & - & - & 1-10 \\
    5 & at least & Comparison & - & - & 1-20 \\
    1-10 & 300 & Literal & - & - & - \\
    1-20 & 3.0 & Literal & - & - & - \\
    1-13 & height & Property & 1-15 & 1-20 & - \\
    1-5 & capacity & Property & 1-2 & 1-10 & - \\
    1-1 & For & Relation & 1-15 & 1-2 & - \\
    4 & must be & Only & - & - & 1-13 \\
    \bottomrule
\end{tabular}
\end{subtable}
\end{minipage}
\end{table}

\begin{figure}[htb!]
\begin{subfigure}[htb!]{0.475\linewidth}
    \centering
    \includegraphics[width=\linewidth]{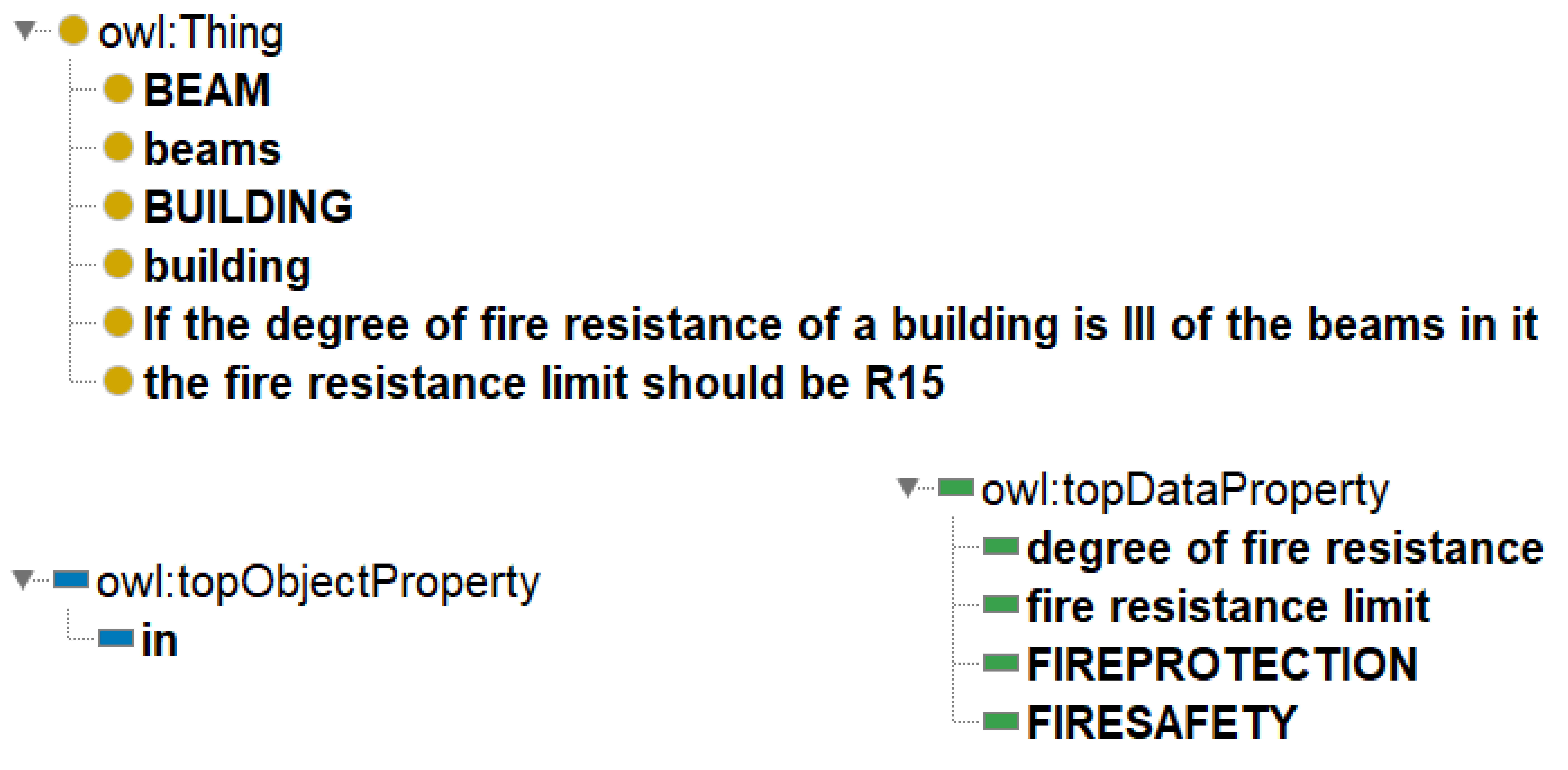}
    \caption{Based on \nameref{ex1}}
    \label{fig:ex1_entities}
\end{subfigure}
\begin{subfigure}[htb!]{0.475\linewidth}
    \centering
    \includegraphics[width=\linewidth]{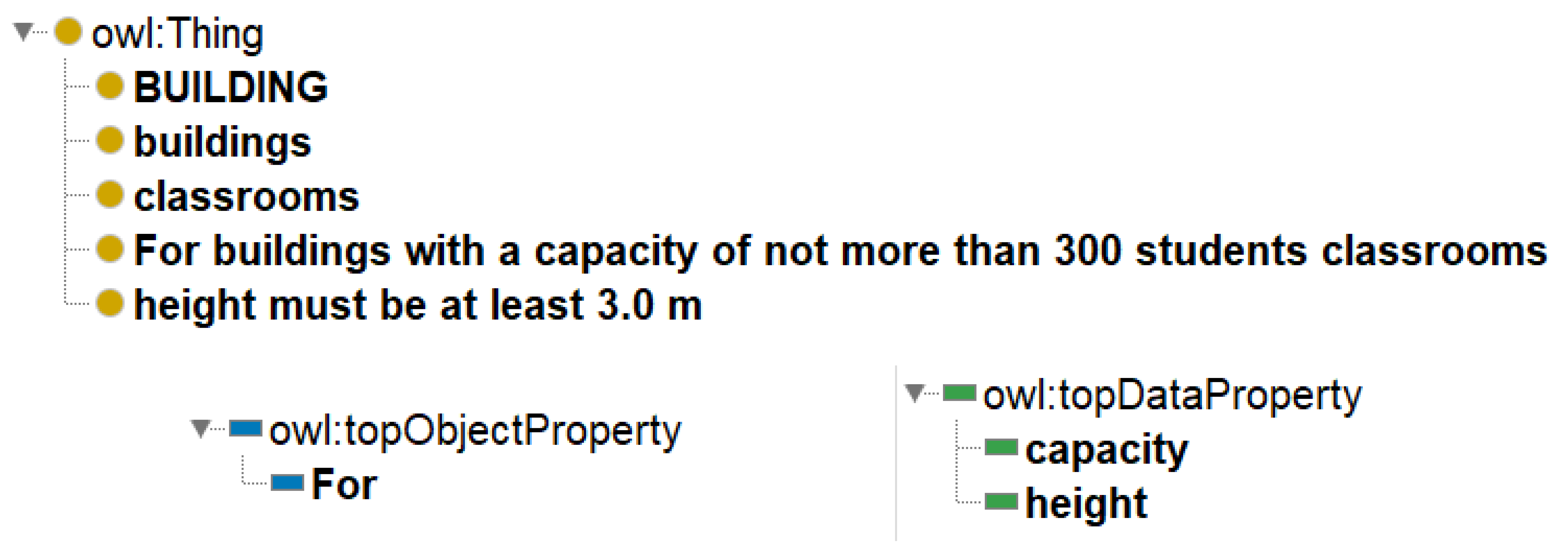}
    \caption{Based on \nameref{ex2}}
    \label{fig:ex2_entities}
\end{subfigure}
\caption{Generated OWL entities}
\end{figure}

\paragraph{Generating entities.}
To transform an annotated regulation into OWL Dl code, we start with an ontology $O$ that includes entities imported from a domain ontology corresponding to $Tr$ (in our use-case, ifcOWL). The first step involves generating elementary entities in $O$ based on the tags \textit{Class}, \textit{Relation}, and \textit{Property} from $ST$, according to the mappings in \autoref{tab:types}. Additionally, OWL classes are created based on the tags \textit{Subject} and \textit{Restriction} from $SR$. The original tokens from the regulation text are assigned as labels for the created entities. \autoref{fig:ex1_entities} demonstrates OWL entities generated based on \nameref{ex1} and \autoref{fig:ex2_entities} for \nameref{ex2}.

\paragraph{Entity alignment.}
Once the elementary entities are created, we establish connections between them across different annotation layers based on token inclusion. Specifically, for each Term tag, we check whether its tokens also belong to any \textit{Class} tag. If so, we add the \texttt{owl:equivalentTo} relation. These connections ensure that domain objects align with regulation classes and are further classified through class restrictions into regulation subjects. In \nameref{ex1}, the equivalence between the \texttt{ifc:IfcBeam} class and the \texttt{beams} OWL class is generated. This guarantees that any object identified as belonging to the \textit{ifc:IfcBeam} class in a BIM model also belongs to the \texttt{beams} class in the ontology. The same for the \texttt{ifc:IfcBuilding} and \texttt{buildings} classes.

\paragraph{Vocabulary mapping.}
To balance annotation complexity with ontology generation accuracy, we employ vocabulary-based mappings for certain tokens from $ST$ to OWL operators. Specifically, we define two mappings: $Card: T\rightarrow N$, which matches strings with integers $N$ for constructing cardinal restrictions, and $Constr: T\rightarrow\{\leq,=,\geq\}$, which is used to define constrained data types. For instance, in \nameref{ex2}, the token ``not more than'', labeled with the tag \textit{Comparison}, is mapped through $Constr$ to \texttt{xsd:maxInclusive}. While no relevant case for the $Card$ mapping is present in our examples, it could be used to map, for instance, the word ``two'' to the integer 2.

\paragraph{Constructing class restrictions.}
Next, we use the remaining tags from $ST$ to construct restricted classes based on the previously generated elementary OWL entities. For shortness, we refer to the combined set of ObjectProperties and DataProperties as predicates $P$. First, we identify the set of classes and literals, denoted as $R$, that appear at the end of predicate chains, i.e. those that are not domains for other predicates. We then apply backward induction to iteratively construct intermediate restricted classes utilizing the $ST$ table. In each iteration, we check if any \textit{Not} or \textit{Or} tags are associated with elements in $R$ and apply \texttt{owl:complementOf} or \texttt{owl:unionOf} respectively. For each $r\in R$ (a class if it corresponds to a \textit{Relation} or a literal if it corresonpds to a \textit{Property}), we identify its incoming predicates $P$ through $ST$ and determine for every predicate its associated restriction. Predicate restrictions are generated using the tags \textit{Some}, \textit{Only} or \textit{Number}. If a predicate has no annotated restriction, we assign \textit{Some} by default if it belongs to the subject of the regulation and \textit{Only} if it belongs to the requirement. The $Card$ mapping is used to define cardinal restrictions. If a \textit{Literal} is accompanied by a \textit{Comparison} tag, the mapping $Constr$ is used to generate a constrained data type. Each intermediate restricted class $c_{restr}$ is added to the resulting set $C_{restr}$, while the original $r$ is removed from $R$. These intermediate restricted classes are then passed to the next iteration of the algorithm. Once all predicate chains are processed, the terminal restricted classes along with any remaining classes in $R$, i.e. those that are not ranges of any predicates but possibly with complements or unions, form the complete set of the building blocks for defining the regulation's subject and requirement. \autoref{alg:chains} outlines the recursive construction of a set of intermediate restricted classes.

\begin{figure}[htb!]
\centering
\begin{subfigure}[htb!]{0.475\linewidth}
    \includegraphics[width=\linewidth]{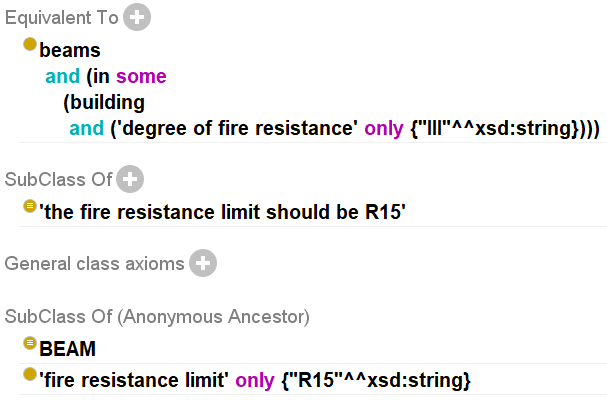}
    \caption{From \nameref{ex1}}
    \label{fig:ex1_axiom}
\end{subfigure}%
\hfill
\begin{subfigure}[htb!]{0.475\linewidth}
    \includegraphics[width=\linewidth]{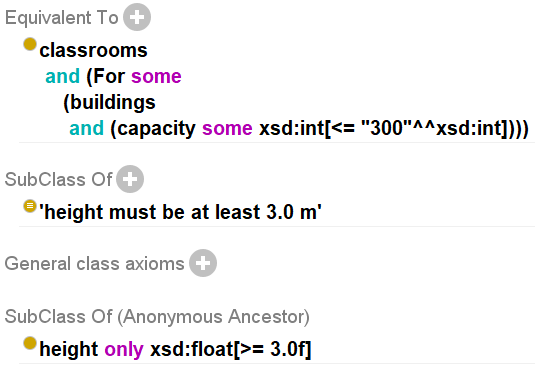}
    \caption{From \nameref{ex2}}
    \label{fig:ex2_axiom}
\end{subfigure}
\caption{Constructed axioms}
\end{figure}

\paragraph{Constructing axioms.}
Finally, using all intermediate classes $C_{restr}$ obtained from \autoref{alg:chains}, we construct class restrictions that accurately capture the semantics of regulation subjects and requirements. This enables domain object classification and compliance checking with OWL reasoning. To achieve this, for each $sr\in SR$, we identify the corresponding class restrictions $C_{sr}\subset C_{restr}$ based on textual token inclusion. The selected classes from $C_{sr}$ are then combined using the \texttt{owl:intersectionOf} relation to form a single complex class $c_{sr}$, which is declared equivalent (\texttt{owl:equivalenTo}) to the given $sr$. Finally, we utilize the \textit{To} arrow between semantic roles in $SR$ to establish an \texttt{owl:subClassOf} relation between the class $s$, which represents the subject of the regulation, with the class $r$, which represents the requirement of the regulation. \autoref{alg:expressions} details this final step. \autoref{fig:ex1_axiom} and \autoref{fig:ex2_axiom} demonstrate the resulting axioms, which represent the semantics of the regulations from \nameref{ex1} and \nameref{ex2}, respectively.


\begin{minipage}{0.475\linewidth}
\begin{algorithm}[H]
\caption{Constructing class restrictions}
\label{alg:chains}
\begin{algorithmic}
\Require $ST$, $R$
\State $R\gets checkOr(R)$
\State $R\gets checkNot(R)$
\State $C_{restr}\gets \emptyset$
\For{$r\in R$}
    \State $P\gets ST(r)$
    \For{$p\in P$}
        \State $restr\gets ST(P)$
        \State $c_{restr}\gets restrictedClass(p, r, restr)$
        \State $C_{restr}\gets C_{restr}\cup \{c_{restr}\}$
    \EndFor
    \State $R\gets R\setminus\{r\}$
\EndFor
\State Apply \autoref{alg:chains} to $ST$, $C_{restr}$
\State $C_{restr}\gets C_{restr}\cup R$
\end{algorithmic}
\end{algorithm}
\end{minipage}
\hfill
\begin{minipage}{0.475\linewidth}
\begin{algorithm}[H]
\caption{Constructing axioms}
\label{alg:expressions}
\begin{algorithmic}
\Require $SR$, $C_{restr}$
\For{$sr\in SR$}
    \State $C_{sr}\gets select(C_{restr}, sr)$
    \State $c_{sr} \gets \bigcap C_{sr}$
    \State $sr\equiv c_{sr}$
\EndFor
\For{$To\in SR$}
    \State $(s, r)\gets SR(To)$
    \State $s\subseteq r$
\EndFor
\end{algorithmic}
\end{algorithm}
\end{minipage}

As a result, given an annotated regulation, the proposed method, with certain limitations, generates an ontology in machine-interpretable OWL DL code. In this ontology, domain terms are connected to restricted classes representing the subject and requirement of the regulation. Finally, subjects and requirements are connected using the \texttt{owl:subClassOf} relation. Consequently, the ontology enables the classification of objects described by domain terms into regulation subjects and facilitates trustworthy compliance checking through OWL reasoning.
\section{Proof of Concept}
\label{sec:proof}

To validate the proposed approach, we developed a proof of concept that converts annotated regulations exported from INCEpTIONs into OWL DL code using the described algorithm. We applied this prototype to the regulations from \nameref{ex1} and \nameref{ex2} and validated the resulting OWL DL representations through a compliance-checking scenario with modeled data.



\paragraph{Prototype.}
The prototype receives annotated regulations exported from INCEpTION in WebAnno TSV v3.3 format and generates OWL DL code with machine-actionable regulations following the proposed algorithm. The prototype is implemented in Python using the Owlready2 library \cite{Owlready}.

\paragraph{Data.}
\label{data}
To validate the OWL code generated based on \nameref{ex1}, we create six ``closed'' individuals: building (\autoref{lst:building}), building with the degree of fire resistance III (\autoref{lst:building_III}), beam (\autoref{lst:beam}), beam in the building (\autoref{lst:beam_in}), beam in the building of the degree of fire resistance III (\autoref{lst:beam_in_III}), and beam in the building of the degree of fire resistance III with fire resistance limit R15 (\autoref{lst:beam_in_III_R15}). According to the complexity levels defined in \cite{bonduel2018ifc}, the last one is classified as level L3, as evaluating this individual involves three triples: \textit{(beam, FIREPROTECTION, R15)}, \textit{(beam, in, building\_III)} and \textit{(building, FIRESAFETY, III)}.

\begin{minipage}{.475\linewidth}
\begin{lstlisting}[label=lst:building, caption={Building}, breaklines=true, basicstyle=\ttfamily\small]
Individual: building

    Types: 
        BUILDING,
        in only owl:Nothing
\end{lstlisting}
\end{minipage}%
\hfill
\begin{minipage}{.475\linewidth}
\begin{lstlisting}[label=lst:beam, caption={Beam}, breaklines=true, basicstyle=\ttfamily\small]
Individual: beam

    Types: 
        BEAM,
        in only owl:Nothing
\end{lstlisting}
\end{minipage}

\paragraph{}

\begin{minipage}{.475\linewidth}
\begin{lstlisting}[label=lst:building_III, caption={Building of the degree of fire resistance III}, breaklines=true, basicstyle=\ttfamily\small]
Individual: building_III

    Types: 
        BUILDING,
        in only owl:Nothing,
        FIRESAFETY only {"III"^^xsd:string}
    
    Facts:  
     FIRESAFETY  "III"^^xsd:string
\end{lstlisting}
\end{minipage}%
\hfill
\begin{minipage}{.475\linewidth}
\begin{lstlisting}[label=lst:beam_in, caption={Beam in the building}, breaklines=true, basicstyle=\ttfamily\small]
Individual: beam_in

    Types: 
        BEAM,
        in only ({building})
    
    Facts:  
     in  building
\end{lstlisting}
\end{minipage}

\begin{minipage}{.475\linewidth}
\begin{lstlisting}[label=lst:beam_in_III, caption={Beam in the building of the degree of fire resistance III}, breaklines=true, basicstyle=\ttfamily\small]
Individual: beam_in_III

    Types: 
        BEAM,
        in only ({building_III})
    
    Facts:  
     in  building_III
\end{lstlisting}
\end{minipage}%
\hfill
\begin{minipage}{.475\linewidth}
\begin{lstlisting}[label=lst:beam_in_III_R15, caption={Beam in the building of the degree of fire resistance III with fire resistance limit R15}, breaklines=true, basicstyle=\ttfamily\small]
Individual: beam_in_III_R15

    Types: 
        BEAM,
        in only ({building_III})
    
    Facts:  
     in  building_III,
     FIREPROTECTION  "R 15"^^xsd:string
\end{lstlisting}
\end{minipage}

\paragraph{Validation script.}
To validate the generated OWL DL code, a user scenario was developed. It consists of the following steps:
\begin{enumerate}
    \item Choose a regulation $r$.
    \item Generate onto $O$ from $r$ using the prototype.
    \item Import individuals representing domain objects that comply with $r$.
    \item Run reasoner.
    \item Modify individuals so that they do not comply to $r$.
    \item Run reasoner.
\end{enumerate}
If the generated OWL DL code is correct, the reasoning in Step 4 should successfully classify the individuals as subjects of $r$. However, after Step 6, the reasoner is expected to detect noncompliance as $O$ becomes inconsistent. This scenario was implemented in Python using the Owlready2 library and Pellet reasoner \cite{Pellet}.

\paragraph{Results.}
We applied the described scenario to the knowledge graph obtained by importing the individuals described in \nameref{data} into the ontology generated from \nameref{ex1}. The logs from the initial reasoner run are provided in \autoref{lst:reasoning1}. As expected, the process completed successfully in 3.28 sec. Among other results, it correctly classified the instances \texttt{beam}, \texttt{beam\_in\_III}, and \texttt{beam\_in\_III\_R15} as subjects of the regulation. Finally, we modified the fire resistance limit of \texttt{beam\_in\_III\_R15} from ``R15'' to `''R14'', and reran the reasoner. As expected, it raised an error due to ontology inconsistency. \autoref{lst:explanation} Provides Pellet's explanation for this inconsistency. These results confirm the validity of the generated OWL DL code.

\begin{lstlisting}[label=lst:reasoning1, float=htb!, caption={Successful reasoning}, breaklines=true, basicstyle=\ttfamily\small]
* Owlready2 * Running Pellet...
* Owlready2 * Pellet took 3.277837038040161 seconds
* Owlready2 * Pellet output:
...
* Owlready * Reparenting reg.beam: {reg.BEAM} => {reg.Subject}
* Owlready * Reparenting reg.beam_in: {reg.BEAM} => {reg.Subject, reg.BEAM}
* Owlready * Reparenting reg.beam_in_III: {reg.BEAM} => {reg.Subject}
* Owlready * Reparenting reg.beam_in_III_R15: {reg.BEAM} => {reg.Subject}
...
\end{lstlisting}

\begin{lstlisting}[label=lst:explanation, float=htb!, caption={Explanation of inconsistency}, breaklines=true, basicstyle=\ttfamily\small]
Explanation for: owl:Thing SubClassOf owl:Nothing
1) EquivalentProperties: FIREPROTECTION, 'fire resistance limit'
2) 'beam_in_III_R15' Type 'If the degree of fire resistance of a building is III the beams in it'
3) 'the fire resistance limit should be R15' EquivalentTo 'fire resistance limit' only {"R15"^^xsd:string}
4) 'If the degree of fire resistance of a building is III the beams in it' SubClassOf 'the fire resistance limit should be R15'
5) beam_in_III FIREPROTECTION "R14"^^xsd:string
\end{lstlisting}

%
\section{Conclusion}
\label{sec:conc}

In this study, we addressed the challenge of converting normative regulations into a machine-interpretable OWL DL code to enable automatic compliance checking using general-purpose reasoning. To facilitate the formalization, we proposed a semantic annotation schema for regulation texts that includes three tag layers: domain Terms, Semantic Types and Semantic Roles, and a number of arrows between the tags. Additionally, we developed an algorithm to convert annotated regulations into OWL DL code. In the resulting OWL DL representations, domain terms are connected to restricted OWL classes, and the requirements expressed in the regulations are represented as general class inclusion axioms. To validate the proposed method, we implemented a proof of concept and a validation script. The prototype was successfully applied to examples from the building construction domain.

\paragraph{Limitations.} The proposed approach has several limitations. First, it relies on vocabulary mappings to generate cardinality restrictions and restricted data types. While the set of words representing natural numbers or comparisons is countable, any new ones or typos in regulatory texts require manual processing. Second, not all relational restrictions or logical connectives are explicitly stated in natural language text. Our approach assigns default values in such cases, but this can potentially lead to inaccuracies in regulation modeling. Specifically, in \nameref{ex1}, it can be stated that the fire resistance limit of the beams should be \textit{at least} R15, rather than strictly R15. If this is the case, the automated generation of the corresponding OWL code from the annotation becomes infeasible, as the comparison is not explicitly mentioned in the text. However, with human intervention, this interpretation can still be formalized in OWL DL as \texttt{'fire resistance limit' \textbf{only} xsd:float[>= 15.0f]}. Finally, no ontology is entirely comprehensive for any application domain. For instance, the regulation in \nameref{ex2} applies to classrooms, yet the ifcOWL ontology lacks a dedicated class for classrooms. This gap between domain data and regulations must be addressed to enable automated compliance checking.  

\paragraph{Future work.} In the future, we will focus on automating regulation annotation using machine learning models. We believe that approaches relying on automatically generating code for ACC or performing direct compliance checking with ML will never be trustworthy enough for full automation. In contrast, our approach will leverage machine learning solely to suggest annotation tags, while the semantic modeling of regulations remains under human control. This ensures that human autonomy and decision-making are preserved, aligning with ethical principles for AI development and deployment, such as the Artificial Intelligence Act\footnote{\url{https://artificialintelligenceact.eu/}} and the EU Ethics Guidelines for Trustworthy AI\footnote{\url{https://digital-strategy.ec.europa.eu/en/library/ethics-guidelines-trustworthy-ai}}.




\section*{Declaration on Generative AI}
During the preparation of this work, the author(s) used ChatGPT in order to: Grammar and spelling check. After using these tool(s)/service(s), the author(s) reviewed and edited the content as needed and take(s) full responsibility for the publication’s content. 
\bibliography{sample-ceur}

\begin{thebibliography}{33}
\expandafter\ifx\csname natexlab\endcsname\relax\def\natexlab#1{#1}\fi
\providecommand{\url}[1]{\texttt{#1}}
\providecommand{\href}[2]{#2}
\providecommand{\path}[1]{#1}
\providecommand{\DOIprefix}{doi:}
\providecommand{\ArXivprefix}{arXiv:}
\providecommand{\URLprefix}{URL: }
\providecommand{\Pubmedprefix}{pmid:}
\providecommand{\doi}[1]{\href{http://dx.doi.org/#1}{\path{#1}}}
\providecommand{\Pubmed}[1]{\href{pmid:#1}{\path{#1}}}
\providecommand{\bibinfo}[2]{#2}
\ifx\xfnm\relax \def\xfnm[#1]{\unskip,\space#1}\fi
\bibitem[{Zhang and El-Gohary(2017)}]{Zhang2017Jan}
\bibinfo{author}{J.~Zhang}, \bibinfo{author}{N.~M. El-Gohary},
\newblock \bibinfo{title}{{Integrating semantic NLP and logic reasoning into a
  unified system for fully-automated code checking}},
\newblock \bibinfo{journal}{Autom. Constr.} \bibinfo{volume}{73}
  (\bibinfo{year}{2017}) \bibinfo{pages}{45--57}.
  \DOIprefix\doi{10.1016/j.autcon.2016.08.027}.
\bibitem[{Fitkau and Hartmann(2024)}]{fitkau2024ontology}
\bibinfo{author}{I.~Fitkau}, \bibinfo{author}{T.~Hartmann},
\newblock \bibinfo{title}{An ontology-based approach of automatic compliance
  checking for structural fire safety requirements},
\newblock \bibinfo{journal}{Advanced Engineering Informatics}
  \bibinfo{volume}{59} (\bibinfo{year}{2024}) \bibinfo{pages}{102314}.
\bibitem[{van Berlo et~al.(2024)van Berlo, Costa, Klooster, Breitenfelder,
  Lavikka, Schneider, and Paasiala}]{vanbim}
\bibinfo{author}{L.~van Berlo}, \bibinfo{author}{G.~Costa},
  \bibinfo{author}{R.~Klooster}, \bibinfo{author}{K.~Breitenfelder},
  \bibinfo{author}{R.~Lavikka}, \bibinfo{author}{K.~Schneider},
  \bibinfo{author}{P.~Paasiala},
\newblock \bibinfo{title}{Bim information reliability consequences for digital
  permit checking},
\newblock \bibinfo{year}{2024}.
\bibitem[{Nuyts et~al.(2024)Nuyts, Bonduel, and
  Verstraeten}]{nuyts2024comparative}
\bibinfo{author}{E.~Nuyts}, \bibinfo{author}{M.~Bonduel},
  \bibinfo{author}{R.~Verstraeten},
\newblock \bibinfo{title}{Comparative analysis of approaches for automated
  compliance checking of construction data},
\newblock \bibinfo{journal}{Advanced Engineering Informatics}
  \bibinfo{volume}{60} (\bibinfo{year}{2024}) \bibinfo{pages}{102443}.
\bibitem[{Donkers and Petrova(2024)}]{donkers2024converting}
\bibinfo{author}{A.~J. Donkers}, \bibinfo{author}{E.~Petrova},
\newblock \bibinfo{title}{Converting fire safety regulations to shacl shapes
  using natural language processing},
\newblock in: \bibinfo{booktitle}{Proceedings of the 3rd NLP4KGC: Natural
  Language Processing for Knowledge Graph Construction co-located with the 20th
  International Conference on Semantic Systems (SEMANTiCS 2024)},
  \bibinfo{organization}{CEUR-WS. org}, \bibinfo{year}{2024}.
\bibitem[{Fuchsa et~al.(2024)Fuchsa, Dimyadia, Witbrocka, and
  Amora}]{fuchsa2024transformer}
\bibinfo{author}{S.~Fuchsa}, \bibinfo{author}{J.~Dimyadia},
  \bibinfo{author}{M.~Witbrocka}, \bibinfo{author}{R.~Amora},
\newblock \bibinfo{title}{Transformer-based semantic parsing of building
  reg-ulations: Towards supporting regulators in drafting machine-readable
  rules},
\newblock in: \bibinfo{booktitle}{Digital Building Permit Conference 2024},
  \bibinfo{year}{2024}.
\bibitem[{Chen et~al.(2024)Chen, Lin, Jiang, and An}]{chen2024automated}
\bibinfo{author}{N.~Chen}, \bibinfo{author}{X.~Lin},
  \bibinfo{author}{H.~Jiang}, \bibinfo{author}{Y.~An},
\newblock \bibinfo{title}{Automated building information modeling compliance
  check through a large language model combined with deep learning and
  ontology},
\newblock \bibinfo{journal}{Buildings} \bibinfo{volume}{14}
  (\bibinfo{year}{2024}) \bibinfo{pages}{1983}.
\bibitem[{Hashmi et~al.(2015)Hashmi, Governatori, and Wynn}]{Hashmi}
\bibinfo{author}{M.~Hashmi}, \bibinfo{author}{G.~Governatori},
  \bibinfo{author}{M.~Wynn},
\newblock \bibinfo{title}{Normative requirements for regulatory compliance: An
  abstract formal framework},
\newblock \bibinfo{journal}{Information Systems Frontiers} \bibinfo{volume}{18}
  (\bibinfo{year}{2015}). \DOIprefix\doi{10.1007/s10796-015-9558-1}.
\bibitem[{Athan et~al.(2015)Athan, Governatori, Palmirani, Paschke, and
  Wyner}]{LegalRuleML}
\bibinfo{author}{T.~Athan}, \bibinfo{author}{G.~Governatori},
  \bibinfo{author}{M.~Palmirani}, \bibinfo{author}{A.~Paschke},
  \bibinfo{author}{A.~Wyner}, \bibinfo{title}{LegalRuleML: Design Principles
  and Foundations}, \bibinfo{publisher}{Springer International Publishing},
  \bibinfo{address}{Cham}, \bibinfo{year}{2015}, pp. \bibinfo{pages}{151--188}.
  \DOIprefix\doi{10.1007/978-3-319-21768-0_6}.
\bibitem[{Gandon et~al.(2017)Gandon, Governatori, and Villata}]{Gandon}
\bibinfo{author}{F.~Gandon}, \bibinfo{author}{G.~Governatori},
  \bibinfo{author}{S.~Villata},
\newblock \bibinfo{title}{Normative requirements as linked data},
\newblock in: \bibinfo{booktitle}{JURIX 2017-The 30th international conference
  on Legal Knowledge and Information Systems}, \bibinfo{year}{2017}, pp.
  \bibinfo{pages}{1--10}. \DOIprefix\doi{10.3233/978-1-61499-838-9-1}.
\bibitem[{Francesconi(2016)}]{legal}
\bibinfo{author}{E.~Francesconi},
\newblock \bibinfo{title}{Semantic model for legal resources: Annotation and
  reasoning over normative provisions},
\newblock \bibinfo{journal}{Semantic Web} \bibinfo{volume}{7}
  (\bibinfo{year}{2016}) \bibinfo{pages}{255--265}.
  \DOIprefix\doi{10.3233/SW-140150}.
\bibitem[{Lam and Hashmi(2018)}]{Lam}
\bibinfo{author}{H.-P. Lam}, \bibinfo{author}{M.~Hashmi},
\newblock \bibinfo{title}{Enabling reasoning with legalruleml},
\newblock \bibinfo{journal}{Theory and Practice of Logic Programming}
  \bibinfo{volume}{19} (\bibinfo{year}{2018}) \bibinfo{pages}{1--26}.
  \DOIprefix\doi{10.1017/S1471068418000339}.
\bibitem[{Yurchyshyna and Zarli(2009)}]{yurchyshyna2009}
\bibinfo{author}{A.~Yurchyshyna}, \bibinfo{author}{A.~Zarli},
\newblock \bibinfo{title}{An ontology-based approach for formalisation and
  semantic organisation of conformance requirements in construction},
\newblock \bibinfo{journal}{Automation in Construction} \bibinfo{volume}{18}
  (\bibinfo{year}{2009}) \bibinfo{pages}{1084--1098}.
\bibitem[{Pauwels et~al.(2011)Pauwels, Van~Deursen, Verstraeten, De~Roo,
  De~Meyer, Van~de Walle, and Van~Campenhout}]{pauwels2011}
\bibinfo{author}{P.~Pauwels}, \bibinfo{author}{D.~Van~Deursen},
  \bibinfo{author}{R.~Verstraeten}, \bibinfo{author}{J.~De~Roo},
  \bibinfo{author}{R.~De~Meyer}, \bibinfo{author}{R.~Van~de Walle},
  \bibinfo{author}{J.~Van~Campenhout},
\newblock \bibinfo{title}{A semantic rule checking environment for building
  performance checking},
\newblock \bibinfo{journal}{Automation in construction} \bibinfo{volume}{20}
  (\bibinfo{year}{2011}) \bibinfo{pages}{506--518}.
\bibitem[{Beach et~al.(2015)Beach, Rezgui, Li, and Kasim}]{beach2015}
\bibinfo{author}{T.~H. Beach}, \bibinfo{author}{Y.~Rezgui},
  \bibinfo{author}{H.~Li}, \bibinfo{author}{T.~Kasim},
\newblock \bibinfo{title}{A rule-based semantic approach for automated
  regulatory compliance in the construction sector},
\newblock \bibinfo{journal}{Expert Systems with Applications}
  \bibinfo{volume}{42} (\bibinfo{year}{2015}) \bibinfo{pages}{5219--5231}.
\bibitem[{Fahad et~al.(2016)Fahad, Bus, and Andrieux}]{Fahad2016Oct}
\bibinfo{author}{M.~Fahad}, \bibinfo{author}{N.~Bus},
  \bibinfo{author}{F.~Andrieux},
\newblock \bibinfo{title}{Towards mapping certification rules over bim},
\newblock in: \bibinfo{booktitle}{CIB W78 Conference},
  volume~\bibinfo{volume}{3}, \bibinfo{year}{2016}.
\bibitem[{Shi and Roman(2017)}]{Shi2017Jan}
\bibinfo{author}{L.~Shi}, \bibinfo{author}{D.~Roman},
\newblock \bibinfo{title}{From standards and regulations to executable rules:
  {A} case study in the building accessibility domain},
\newblock in: \bibinfo{editor}{N.~Bassiliades}, \bibinfo{editor}{A.~Bikakis},
  \bibinfo{editor}{S.~Costantini}, \bibinfo{editor}{E.~Franconi},
  \bibinfo{editor}{A.~Giurca}, \bibinfo{editor}{R.~Kontchakov},
  \bibinfo{editor}{T.~Patkos}, \bibinfo{editor}{F.~Sadri},
  \bibinfo{editor}{W.~V. Woensel} (Eds.), \bibinfo{booktitle}{Proceedings of
  the Doctoral Consortium, Challenge, Industry Track, Tutorials and Posters @
  RuleML+RR 2017 hosted by International Joint Conference on Rules and
  Reasoning 2017 (RuleML+RR 2017), London, UK, July 11-15, 2017}, volume
  \bibinfo{volume}{1875} of \textit{\bibinfo{series}{{CEUR} Workshop
  Proceedings}}, \bibinfo{publisher}{CEUR-WS.org}, \bibinfo{year}{2017}.
\bibitem[{Stolk and McGlinn(2020)}]{stolk2020}
\bibinfo{author}{S.~Stolk}, \bibinfo{author}{K.~McGlinn},
\newblock \bibinfo{title}{Validation of ifcowl datasets using shacl},
\newblock in: \bibinfo{booktitle}{Proceedings of the 8th Linked Data in
  Architecture and Construction Workshop}, \bibinfo{year}{2020}, pp.
  \bibinfo{pages}{91--104}.
\bibitem[{Kovacs and Micsik(2021)}]{kovacs2021bim}
\bibinfo{author}{A.~T. Kovacs}, \bibinfo{author}{A.~Micsik},
\newblock \bibinfo{title}{Bim quality control based on requirement linked
  data},
\newblock \bibinfo{journal}{International Journal of Architectural Computing}
  \bibinfo{volume}{19} (\bibinfo{year}{2021}) \bibinfo{pages}{431--448}.
\bibitem[{Zentgraf et~al.(2022)Zentgraf, Hagedorn, and
  K{\"o}nig}]{zentgraf2022multi}
\bibinfo{author}{S.~Zentgraf}, \bibinfo{author}{P.~Hagedorn},
  \bibinfo{author}{M.~K{\"o}nig},
\newblock \bibinfo{title}{Multi-requirements ontology engineering for automated
  processing of document-based building codes to linked building data
  properties},
\newblock in: \bibinfo{booktitle}{IOP Conference Series: Earth and
  Environmental Science}, volume \bibinfo{volume}{1101},
  \bibinfo{organization}{IOP Publishing}, \bibinfo{year}{2022}, p.
  \bibinfo{pages}{092007}.
\bibitem[{Nuyts et~al.(2023)Nuyts, Werbrouck, Verstraeten, and
  Deprez}]{nuyts2023validation}
\bibinfo{author}{E.~Nuyts}, \bibinfo{author}{J.~Werbrouck},
  \bibinfo{author}{R.~Verstraeten}, \bibinfo{author}{L.~Deprez},
\newblock \bibinfo{title}{Validation of building models against legislation
  using shacl},
\newblock in: \bibinfo{booktitle}{LDAC2023: Linked Data in Architecture and
  Construction Week}, volume \bibinfo{volume}{3633},
  \bibinfo{organization}{CEUR}, \bibinfo{year}{2023}, pp.
  \bibinfo{pages}{164--175}.
\bibitem[{Patlakas et~al.(2024)Patlakas, Christovasilis, Riparbelli, Cheung,
  and Vakaj}]{patlakas2024semantic}
\bibinfo{author}{P.~Patlakas}, \bibinfo{author}{I.~Christovasilis},
  \bibinfo{author}{L.~Riparbelli}, \bibinfo{author}{F.~K. Cheung},
  \bibinfo{author}{E.~Vakaj},
\newblock \bibinfo{title}{Semantic web-based automated compliance checking with
  integration of finite element analysis},
\newblock \bibinfo{journal}{Advanced Engineering Informatics}
  \bibinfo{volume}{61} (\bibinfo{year}{2024}) \bibinfo{pages}{102448}.
\bibitem[{Leinberger et~al.(2020)Leinberger, Seifer, Rienstra, L{\"a}mmel, and
  Staab}]{leinberger2020}
\bibinfo{author}{M.~Leinberger}, \bibinfo{author}{P.~Seifer},
  \bibinfo{author}{T.~Rienstra}, \bibinfo{author}{R.~L{\"a}mmel},
  \bibinfo{author}{S.~Staab},
\newblock \bibinfo{title}{Deciding shacl shape containment through description
  logics reasoning},
\newblock in: \bibinfo{booktitle}{The Semantic Web--ISWC 2020: 19th
  International Semantic Web Conference, Athens, Greece, November 2--6, 2020,
  Proceedings, Part I 19}, \bibinfo{organization}{Springer},
  \bibinfo{year}{2020}, pp. \bibinfo{pages}{366--383}.
\bibitem[{Bogaerts et~al.(2022)Bogaerts, Jakubowski, and Van~den
  Bussche}]{bogaerts2022shacl}
\bibinfo{author}{B.~Bogaerts}, \bibinfo{author}{M.~Jakubowski},
  \bibinfo{author}{J.~Van~den Bussche},
\newblock \bibinfo{title}{Shacl: A description logic in disguise},
\newblock in: \bibinfo{booktitle}{International Conference on Logic Programming
  and Nonmonotonic Reasoning}, \bibinfo{organization}{Springer},
  \bibinfo{year}{2022}, pp. \bibinfo{pages}{75--88}.
\bibitem[{Lamy(2017)}]{Owlready}
\bibinfo{author}{J.-B. Lamy},
\newblock \bibinfo{title}{Owlready: Ontology-oriented programming in python
  with automatic classification and high level constructs for biomedical
  ontologies},
\newblock \bibinfo{journal}{Artificial Intelligence in Medicine}
  \bibinfo{volume}{80} (\bibinfo{year}{2017}) \bibinfo{pages}{11--28}.
  \DOIprefix\doi{https://doi.org/10.1016/j.artmed.2017.07.002}.
\bibitem[{Ahmetaj et~al.(2021)Ahmetaj, David, Ortiz, Polleres, Shehu, and
  Simkus}]{ahmetaj2021reasoning}
\bibinfo{author}{S.~Ahmetaj}, \bibinfo{author}{R.~David},
  \bibinfo{author}{M.~Ortiz}, \bibinfo{author}{A.~Polleres},
  \bibinfo{author}{B.~Shehu}, \bibinfo{author}{M.~Simkus},
\newblock \bibinfo{title}{Reasoning about explanations for non-validation in
  shacl},
\newblock in: \bibinfo{booktitle}{Description Logics}, \bibinfo{year}{2021}.
\bibitem[{Hjelseth and Nisbet(2011)}]{Hjelseth2011}
\bibinfo{author}{E.~Hjelseth}, \bibinfo{author}{N.~N. Nisbet},
\newblock \bibinfo{title}{Capturing normative constraints by use of the
  semantic mark-up rase methodology},
\newblock in: \bibinfo{booktitle}{Proceedings of CIB W78-W102 Conference},
  \bibinfo{year}{2011}, pp. \bibinfo{pages}{1--10}.
\bibitem[{Hettiarachchi et~al.(2025)Hettiarachchi, Dridi, Gaber, Parsafard,
  Bocaneala, Breitenfelder, Costa, Hedblom, Juganaru-Mathieu, Mecharnia
  et~al.}]{hettiarachchi2025code}
\bibinfo{author}{H.~Hettiarachchi}, \bibinfo{author}{A.~Dridi},
  \bibinfo{author}{M.~M. Gaber}, \bibinfo{author}{P.~Parsafard},
  \bibinfo{author}{N.~Bocaneala}, \bibinfo{author}{K.~Breitenfelder},
  \bibinfo{author}{G.~Costa}, \bibinfo{author}{M.~Hedblom},
  \bibinfo{author}{M.~Juganaru-Mathieu}, \bibinfo{author}{T.~Mecharnia},
  et~al.,
\newblock \bibinfo{title}{Code-accord: A corpus of building regulatory data for
  rule generation towards automatic compliance checking},
\newblock \bibinfo{journal}{Scientific Data} \bibinfo{volume}{12}
  (\bibinfo{year}{2025}) \bibinfo{pages}{170}.
\bibitem[{Eastman et~al.(2011)Eastman, Eastman, Teicholz, Sacks, and
  Liston}]{BIM}
\bibinfo{author}{C.~M. Eastman}, \bibinfo{author}{C.~Eastman},
  \bibinfo{author}{P.~Teicholz}, \bibinfo{author}{R.~Sacks},
  \bibinfo{author}{K.~Liston}, \bibinfo{title}{BIM handbook: A guide to
  building information modeling for owners, managers, designers, engineers and
  contractors}, \bibinfo{publisher}{John Wiley \& Sons}, \bibinfo{year}{2011}.
\bibitem[{Pauwels and Terkaj(2016)}]{ifcOWL}
\bibinfo{author}{P.~Pauwels}, \bibinfo{author}{W.~Terkaj},
\newblock \bibinfo{title}{Express to owl for construction industry: Towards a
  recommendable and usable ifcowl ontology},
\newblock \bibinfo{journal}{Automation in Construction} \bibinfo{volume}{63}
  (\bibinfo{year}{2016}) \bibinfo{pages}{100--133}.
  \DOIprefix\doi{https://doi.org/10.1016/j.autcon.2015.12.003}.
\bibitem[{Klie et~al.(2018)Klie, Bugert, Boullosa, de~Castilho, and
  Gurevych}]{Inception}
\bibinfo{author}{J.-C. Klie}, \bibinfo{author}{M.~Bugert},
  \bibinfo{author}{B.~Boullosa}, \bibinfo{author}{R.~E. de~Castilho},
  \bibinfo{author}{I.~Gurevych},
\newblock \bibinfo{title}{The inception platform: Machine-assisted and
  knowledge-oriented interactive annotation},
\newblock in: \bibinfo{booktitle}{Proceedings of the 27th International
  Conference on Computational Linguistics: System Demonstrations},
  \bibinfo{publisher}{Association for Computational Linguistics},
  \bibinfo{year}{2018}, pp. \bibinfo{pages}{5--9}. \bibinfo{note}{Event Title:
  The 27th International Conference on Computational Linguistics (COLING
  2018)}.
\bibitem[{Bonduel et~al.(2018)Bonduel, Oraskari, Pauwels, Vergauwen, and
  Klein}]{bonduel2018ifc}
\bibinfo{author}{M.~Bonduel}, \bibinfo{author}{J.~Oraskari},
  \bibinfo{author}{P.~Pauwels}, \bibinfo{author}{M.~Vergauwen},
  \bibinfo{author}{R.~Klein},
\newblock \bibinfo{title}{The ifc to linked building data converter: current
  status},
\newblock in: \bibinfo{booktitle}{6th International Workshop on Linked Data in
  Architecture and Construction}, \bibinfo{organization}{CEUR-WS. org},
  \bibinfo{year}{2018}, pp. \bibinfo{pages}{34--43}.
\bibitem[{Sirin et~al.(2007)Sirin, Parsia, Grau, Kalyanpur, and Katz}]{Pellet}
\bibinfo{author}{E.~Sirin}, \bibinfo{author}{B.~Parsia}, \bibinfo{author}{B.~C.
  Grau}, \bibinfo{author}{A.~Kalyanpur}, \bibinfo{author}{Y.~Katz},
\newblock \bibinfo{title}{Pellet: A practical owl-dl reasoner},
\newblock \bibinfo{journal}{Web Semant.} \bibinfo{volume}{5}
  (\bibinfo{year}{2007}) \bibinfo{pages}{51–53}.
  \DOIprefix\doi{10.1016/j.websem.2007.03.004}.

\end{thebibliography}

\end{document}